%% file: main.tex
\documentclass[10pt,twocolumn,letterpaper]{article}
\usepackage[accsupp]{axessibility}
\usepackage{cvpr}              

\input{preamble}
\usepackage{mathtools}
\usepackage[most]{tcolorbox}
\usepackage{multirow}
\usepackage{arydshln}

\usepackage[table]{xcolor}
\usepackage{float}

\definecolor{cvprblue}{rgb}{0.21,0.49,0.74}
\usepackage[pagebackref,breaklinks,colorlinks,allcolors=cvprblue]{hyperref}

\def\paperID{11879} 
\def\confName{CVPR}
\def\confYear{2026}

\title{WalkGPT: Grounded Vision–Language Conversation with Depth-Aware Segmentation for Pedestrian Navigation}

\author{
Rafi Ibn Sultan$^{1}$ \quad Hui Zhu$^{1}$ \quad Xiangyu Zhou$^{1}$\\
Chengyin Li$^{2}$ \quad Prashant Khanduri$^{1}$ \quad Marco Brocanelli$^{3}$ \quad Dongxiao Zhu$^{1,4}$\\
$^{1}$\small Department of Computer Science, Wayne State University \quad
$^{2}$\small Department of Radiation Oncology, Henry Ford Health \\
$^{3}$\small Department of Electrical and Computer Engineering, The Ohio State University\quad
$^{4}$\small Institute for AI and Data Science, Wayne State University\\
}
\begin{document}
\maketitle
\input{sec/0_abstract}

\input{sec/1_intro}

\input{sec/2_related_works}
\input{sec/3_methods}

\input{sec/4_experiments}

\input{sec/5_conclusion}

{
    \small
    \bibliographystyle{ieeenat_fullname}
    \bibliography{main}
}

\input{sec/X_suppl}

\end{document}

%% file: preamble.tex
\definecolor{WalkPink}{RGB}{233,30,99}



%% file: sec/0_abstract.tex
\begin{abstract}
Ensuring accessible pedestrian navigation requires reasoning about both semantic and spatial aspects of complex urban scenes, a challenge that existing Large Vision–Language Models (LVLMs) struggle to meet. Although these models can describe visual content, their lack of explicit grounding leads to object hallucinations and unreliable depth reasoning, limiting their usefulness for accessibility guidance. We introduce \textbf{WalkGPT}, a pixel-grounded LVLM for the new task of \textit{Grounded Navigation Guide}, unifying language reasoning and segmentation within a single architecture for depth-aware accessibility guidance. Given a pedestrian-view image and a navigation query, WalkGPT generates a conversational response with segmentation masks that delineate accessible and harmful features, along with relative depth estimation. The model incorporates a Multi-Scale Query Projector (MSQP) that shapes the final image tokens by aggregating them along text tokens across spatial hierarchies, and a Calibrated Text Projector (CTP), guided by a proposed Region Alignment Loss, that maps language embeddings into segmentation-aware representations. These components enable fine-grained grounding and depth inference without user-provided cues or anchor points, allowing the model to generate complete and realistic navigation guidance. We also introduce \textbf{PAVE}, a large-scale benchmark of 41k pedestrian-view images paired with accessibility-aware questions and depth-grounded answers. Experiments show that WalkGPT achieves strong grounded reasoning and segmentation performance. The source code and dataset are available on the \href{https://sites.google.com/view/walkgpt-26/home}{project website}.
\end{abstract}

%% file: sec/1_intro.tex
\section{Introduction}
\label{sec:intro}

Safe and accessible pedestrian navigation is essential for independent mobility in diverse environments. However, pedestrian routes often contain static and dynamic barriers such as stairs, uneven terrain, parked vehicles, and temporary obstructions~\cite{sultan2023geosam,froehlich2025streetviewai,li2024streetviewllm}, posing significant risks for individuals with mobility challenges~\cite{ying2025mmwalk}. While most automated navigation systems are designed for vehicles~\cite{jiang2025survey,zhou2025autovla}, pedestrian-level navigation remains largely underexplored. Addressing this gap requires systems that understand the environment from a first-person view and reason about geometry, depth, and accessibility to navigate complex urban scenes.

\begin{figure}[t]
    \centering
    \includegraphics[width=1.0\columnwidth]{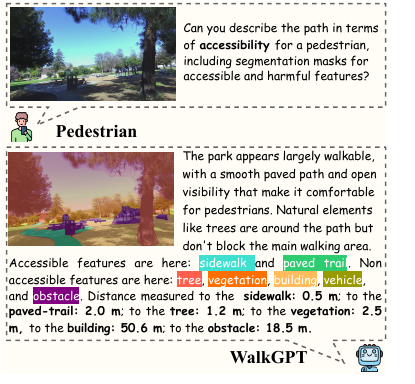}
    \vspace{-9mm}
    \caption{
        Overview of WalkGPT for \textit{accessibility-aware} grounded navigation guide. 
        The model grounds language on segmentation masks enriched with depth information,  
        providing holistic spatial understanding that captures both object shapes and depth cues 
        for interpretable \textit{accessibility analysis}.
        }
    \label{fig:idea}
    \vspace{-5mm}
\end{figure}

Recent advances in Large Vision–Language Models (LVLMs)~\cite{liu2023visual,liu2024improved,dai2023instructblip,li2023blip} demonstrate strong visual understanding and language reasoning capabilities, suggesting potential for conversational guidance in pedestrian navigation. However, existing LVLMs lack explicit spatial reasoning~\cite{chen2024spatialvlm,anonymous2025spatialab,chen2025spatial} needed to infer geometric structure and depth relationships in real-world scenes. Recent spatial-aware approaches~\cite{cai2025depthlm,chen2024spatialvlm,cheng2024spatialrgpt} attempt to address this but often rely on \textit{user-provided visual cues} to estimate object depth, which is impractical for pedestrian navigation because users cannot manually supply such inputs. LVLMs also tend to hallucinate, describing objects not present in the scene~\cite{zhaomitigating}, which can lead to hazardous guidance.

Grounded LVLMs~\cite{rasheed2024glamm,lai2024lisa,zhang2024omg,zhang2024psalm,ren2024pixellm} extend conventional LVLMs by aligning textual references with corresponding image regions, improving spatial consistency and reducing hallucinations~\cite{zhaomitigating}. Their segmentation masks provide explicit shape cues, making them promising for pedestrian navigation. However, these masks are 2D and lack the depth information required for spatial reasoning. Consequently, grounded LVLMs remain limited in understanding relative distances and spatial hierarchy, which are critical for safe, accessibility-aware navigation. Thus, existing grounded LVLMs have not been adapted for pedestrian navigation, a domain challenged by complex urban scenes and the absence of large-scale pedestrian-view datasets with joint question–answering and grounding annotations. These limitations motivate the research question: \textit{Can we develop a grounded LVLM capable of providing depth-aware navigation guidance for pedestrians?}

To address these limitations, we propose \textbf{WalkGPT}, a grounded LVLM with spatial reasoning for the task of \textit{Grounded Navigation Guide}. As illustrated in \Cref{fig:idea}, given a pedestrian-view image and a user query about the path ahead, WalkGPT generates a grounded response integrating free-form reasoning with depth-aware segmentation masks for fine-grained accessibility guidance. Unlike prior spatial reasoning approaches~\cite{cheng2024spatialrgpt,cai2025depthlm}, it operates without requiring user-provided visual cues or anchor points. By grounding accessible and harmful elements along with their segmentation masks and relative distances, WalkGPT provides comprehensive navigation feedback that supports safe movement, including for individuals with disabilities.

WalkGPT, a unified architecture (illustrated in \Cref{fig:architecture}), enables grounded conversations by reasoning over spatial cues from text and segmentation for navigation understanding. It introduces two novel modules: the Multi-Scale Query Projector (MSQP), which aggregates multi-scale visual context, and the Calibrated Text Projector (CTP), guided by a proposed Region Alignment Loss to refine language-to-vision grounding. Using structured tokens such as \texttt{<p>}, \texttt{<assessment>}, \texttt{<SEG>}, and \texttt{<distance>}, WalkGPT structures its outputs to explicitly link language with segmentation and distance cues, enabling automated depth-aware navigation guidance without user-provided inputs. As no existing dataset addressed this problem, we introduce \textbf{PAVE}, a large-scale VQA dataset of 41k pedestrian-view images with accessibility-aware questions and answers containing embedded depth information. 

Our main contributions are summarized as follows:  

\begin{itemize}
    \item We introduce \textbf{WalkGPT}, the first-of-its-kind LVLM for pedestrian accessibility via grounded spatial reasoning.
    \item We propose a streamlined architecture aligning visual and language representations to enhance pixel-level grounding using the novel MSQP and CTP with structured token supervision.
    \item We curate \textbf{PAVE}, a large-scale VQA dataset with depth annotations for accessibility and spatial understanding.
    \item WalkGPT achieves state-of-the-art performance on grounded navigation guidance and sets a benchmark for accessibility-aware AI in pedestrian environments.
\end{itemize}

%% file: sec/2_related_works.tex
\section{Related Works}
\label{sec:related_works}


\noindent \textbf{Grounded Large Vision–Language Models (LVLMs).}  
LVLMs~\cite{alayrac2022flamingo,li2023blip,liu2023visual,dai2023instructblip,bai2023qwen} have advanced multimodal understanding by integrating large language models (LLMs) with visual encoders through large-scale pretraining and instruction tuning. In particular, pixel-grounded LVLMs enable region-level understanding where textual descriptions are explicitly linked to image pixels. Early works performed coarse grounding via bounding boxes~\cite{you2023ferret,chen2023shikra,peng2024grounding,xuan2024pink,chen2024lion,kuckreja2024geochat,ma2024groma,zhang2024llava}. More fine-grained grounding using integrated segmentation encoder–decoder architectures has followed two directions: reasoning segmentation~\cite{lai2024lisa,ren2024pixellm,zhang2024psalm,wang2024llm,yuan2025sa2va} and visually grounded conversation~\cite{zhang2024omg,rasheed2024glamm,xia2024gsva,wu2025f}, both aiming to unify semantic reasoning with pixel-level grounding. Beyond these general-domain efforts, domain-specific studies have adapted visually grounded models to medical~\cite{chen2025mimo} and remote-sensing imagery \cite{shabbir2025geopixel}. However, grounded LVLMs have not yet been explored for accessibility-aware pedestrian navigation, which requires spatial reasoning in complex real-world environments.

\noindent\textbf{Spatial Reasoning LVLMs.}
LVLMs~\cite{cheng2024spatialrgpt,chen2024spatialvlm,cai2025depthlm,gholami2025spatial} designed for spatial reasoning typically use depth maps or visual anchors to capture spatial relations between objects, whereas others~\cite{ning2025enhancing} handle spatial reasoning without performing depth estimation. Although effective for general spatial understanding, these models often depend on user-provided visual cues or anchor points and, in some cases, omit depth prediction, which limits their suitability for navigation-oriented scenarios.

\begin{figure*}[t]
    \centering
    \includegraphics[width=1.0\linewidth]{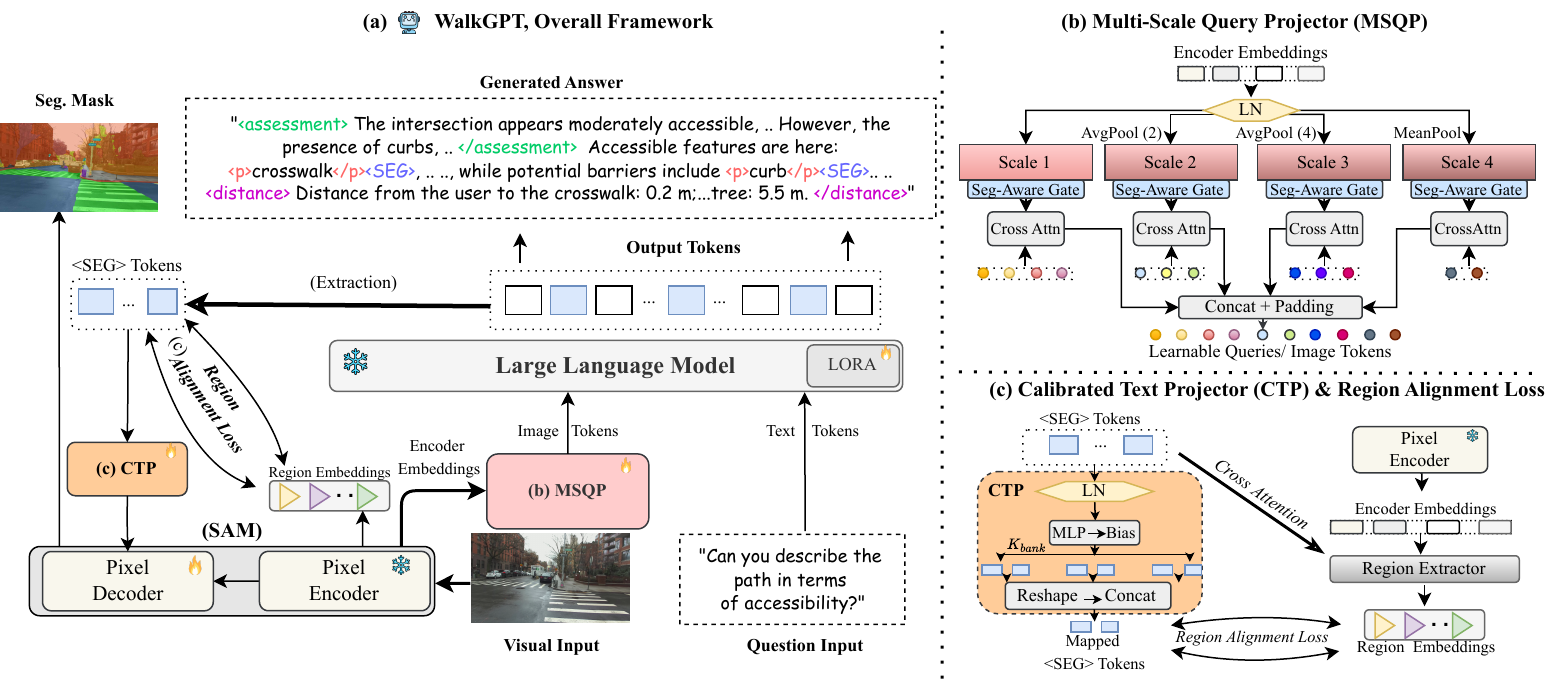}
    \caption{
        Overview of \textbf{WalkGPT} for grounded navigation guidance. 
        (a) Overall framework. 
        (b) The \textbf{Multi-Scale Query Projector (MSQP)}, which aggregates multi-level visual features into spatially aligned image tokens for language reasoning. 
        (c) The \textbf{Calibrated Text Projector (CTP)}, guided by the proposed \textbf{Region Alignment Loss}, maps \texttt{<SEG>} tokens into the visual space. 
        Structured tokens (\texttt{<SEG>}, \texttt{<distance>}, \texttt{<assessment>}, \texttt{<p>}) link language generation with segmentation and depth reasoning.
        }
        \vspace{-2mm}
    \label{fig:architecture}
\end{figure*}

\noindent\textbf{Accessibility-Aware Pedestrian Navigation.}
Basic pedestrian navigation studies~\cite{abbott2018walknet,saha2019project,jiang2022automatic,hwang2024safe,zhou2024navgpt} focus on static object detection or scene labeling with limited insight into route accessibility. More recent LVLM-based or multimodal systems~\cite{froehlich2025streetviewai,li2024streetviewllm,ying2025mmwalk,yuan2024walkvlm} address this gap, but rely on synthetic or metadata-heavy inputs and still lack the fine-grained grounding needed for real-world guidance. Consequently, grounded LVLMs have yet to be explored for accessibility-aware navigation, which requires both pixel-level grounding and depth reasoning in complex urban environments.



%% file: sec/3_methods.tex
\section{Methods}
\subsection{WalkGPT: The Architecture}
\label{sec:methods}
\Cref{fig:architecture} illustrates WalkGPT, a grounded conversational model that generates navigation-aware responses linked to segmentation masks. Text tokens are concatenated with image tokens from a shared SAM-based pixel encoder~\cite{kirillov2023segment} and processed by the LLM to produce the output sequence. Any \texttt{<SEG>} tokens in the response are passed to the SAM pixel decoder to obtain spatially aligned masks. WalkGPT introduces two architectural components, MSQP and CTP, enabling a shared encoder to support both conversation generation and segmentation.

\noindent\textbf{Multi-Scale Query Projector (MSQP).}
The MSQP maps pixel encoder embeddings into semantically aligned image tokens in the language space for LLM input (\Cref{fig:architecture}b). Unlike standard MLP projectors, MSQP aggregates visual features across multiple spatial levels, preserving both local detail and global structure and producing more spatially informative representations.


Given encoder embeddings $\mathbf{Z} \in \mathbb{R}^{B \times L \times C}$ from the pixel encoder, where $B$ is the batch size, $L=H\!\times\!W$ flattened image tokens, and $C$ the feature dimension, we apply a linear projection $\mathbf{W}_{\text{proj}} \in \mathbb{R}^{C \times d_{\text{proj}}}$ to obtain $\mathbf{F}=\mathbf{Z}\mathbf{W}_{\text{proj}} \in \mathbb{R}^{B \times L \times d_{\text{proj}}}$, with $d_{\text{proj}}=1024$ as the working dimension. Then $\mathbf{F}$ is reshaped into a grid and average-pooled at multiple scales, generating token banks $\{\mathbf{x}^{1}, \mathbf{x}^{2}, \mathbf{x}^{4}, \mathbf{x}^{g}\}$ corresponding to native, pooled-by-2, pooled-by-4, and global-mean resolutions. Each scale-specific feature bank is modulated by a segmentation-aware gating function $g(\cdot)$ that highlights structure- and edge-rich regions before attention. 
For each scale $s$, a small set of learnable query embeddings $\mathbf{Q}^{s} \in \mathbb{R}^{Q_s \times d_{\text{proj}}}$ interacts with the gated tokens $\mathbf{x}^{s}$ through two layers of cross-attention, producing refined outputs $\mathbf{O}^{s} \in \mathbb{R}^{B \times Q_s \times d_{\text{proj}}}$. 
Each output token $\mathbf{o}^{s}_i$ is computed as a content-weighted mixture $\mathbf{o}^{s}_i = \sum_{j} \alpha_{ij} \mathbf{x}^{s}_j$, where attention weights $\alpha_{ij}$ are obtained by softmax normalization over query-key similarities. The outputs from all scales are first concatenated and padded to a fixed length $Q=\sum_s Q_s + 4 = 36$ to ensure consistency across scales. The padded sequence is then linearly projected to the LLM hidden dimension $H$, producing the final image tokens $\mathbf{V}_{\text{proj}} \in \mathbb{R}^{B \times Q \times H}$. By attending across multiple spatial hierarchies, MSQP condenses fine-grained details and global scene context into a compact set of tokens $\mathbf{V}_{\text{proj}}$.




\noindent\textbf{Calibrated Text Projector (CTP).}
In our framework, the grounded \texttt{<SEG>} tokens in the generated response (see \Cref{fig:architecture}a) act as textual prompts that guide the pixel decoder for mask prediction, linking language reasoning with pixel-level segmentation. Unlike prior methods that use a linear projection into the segmentation space, CTP expands each token into structured sub-embeddings, preserving fine-grained semantics. These calibrated embeddings improve spatial correspondence, supporting segmentation-aware mask generation.

Given the hidden states of the \texttt{<SEG>} tokens $\mathbf{T} \in \mathbb{R}^{B \times M \times H}$ from the LLM, where $M$ is the number of \texttt{<SEG>} tokens, we apply a linear projection $\mathbf{W}_{\text{vis}} \in \mathbb{R}^{H \times d_{\text{vis}}}$ to transform them, yielding $\mathbf{U} = \mathbf{T}\mathbf{W}_{\text{vis}} \in \mathbb{R}^{B \times M \times d_{\text{vis}}}$, where $d_{\text{vis}}{=}256$ matches the visual backbone dimension. As illustrated in \Cref{fig:architecture}c, to prevent loss of token diversity, each reduced vector $\mathbf{u}_i$ is expanded into a small set of $K_{\text{bank}}$ calibrated embeddings through a bias-augmented transformation $\mathbf{E}_i = \text{MLP}(\mathbf{u}_i) + \mathbf{B}$, where $\mathbf{E}_i \in \mathbb{R}^{K_{\text{bank}} \times d_{\text{vis}}}$ and $\mathbf{B} \in \mathbb{R}^{K_{\text{bank}} \times d_{\text{vis}}}$ contains learnable biases that encode modality-specific priors such as objectness and boundary structure. This expansion allows each token to generate multiple complementary sub-embeddings that capture different aspects of its semantics. The resulting calibrated bank $\mathbf{E} \in \mathbb{R}^{B \times M \times K_{\text{bank}} \times d_{\text{vis}}}$ is reshaped to $\mathbb{R}^{B \times (MK_{\text{bank}}) \times d_{\text{vis}}}$ and concatenated along the token dimension to serve as the text prompt for the pixel decoder.


\noindent\textbf{Region Alignment Loss.}
Since CTP maps the rich $H$-dimensional LLM embeddings (4096) into the lower $d_{\text{vis}}$-dimensional visual space (256), substantial information loss may occur. To preserve semantic detail, we introduce \emph{Region Alignment Loss}, a contrastive regularization that enforces consistency between text embeddings and their corresponding visual regions. For each \texttt{<SEG>} token, the method aligns its projected embedding with the visual features of the target region while pushing it away from unrelated areas, encouraging a semantically faithful $H{\to}d_{\text{vis}}$ mapping. With the pixel encoder frozen, this regularization improves projection fidelity and language–region alignment.

For each sample $b \in \{1, \ldots, B\}$ in the batch, we take the pre-projection \texttt{<SEG>} token embeddings $\mathbf{T}_b \in \mathbb{R}^{M \times H}$ (from the LLM hidden states) and the flattened pixel-encoder embeddings $\mathbf{Z}_b \in \mathbb{R}^{L \times C}$. Since the \texttt{<SEG>} tokens interact with specific image regions during generation, they are used to cross-attend and retrieve the most relevant spatial areas. The attended regions act as pseudo-targets to supervise their projected counterparts from CTP, establishing fine-grained region–token correspondence through cross-attention between the original \texttt{<SEG>} tokens and the visual embeddings. Specifically, we compute
\begin{equation}
\begin{aligned}
\mathbf{q} &= \mathbf{t}_{b,m}\mathbf{W}_q, \quad
\mathbf{K}_b = \mathbf{Z}_b \mathbf{W}_k, \quad
\mathbf{V}_b = \mathbf{Z}_b \mathbf{W}_v, \\
\boldsymbol{\pi} &= \mathrm{softmax}\!\left(\frac{\mathbf{K}_b\mathbf{q}^{\top}}{\sqrt{d_k}}\right)\in\mathbb{R}^{L},
\end{aligned}
\end{equation}
where $\mathbf{t}_{b,m}$ is the $m$-th \texttt{<SEG>} token in batch $b$ and $d_k{=}d_{\text{vis}}$ is the shared query–key projection dimension.

To focus alignment on salient object regions and suppress background noise, we emphasize the top-$K$ image tokens with the highest attention weights. The selected indices are $\mathcal{I}_K = \mathrm{TopK}(\boldsymbol{\pi}, K)$, and the normalized weights are $\alpha_i = \pi_i / \sum_{j \in \mathcal{I}_K}\pi_j$. The positive region embedding is then computed as
\begin{equation}
\mathbf{z}^{+}_{b,m} = \mathbf{W}_o \!\left(\sum_{i \in \mathcal{I}_K} \alpha_i\,\mathbf{v}_{b,i}\right) \in \mathbb{R}^{d_{\text{vis}}},
\end{equation}
where $\mathbf{v}_{b,i}$ denotes the $i$-th value vector from $\mathbf{V}_b$ and full attention is used when $K = L$, i.e., $\mathcal{I}_K$ contains all tokens.

The mapped tokens from CTP, $\mathbf{E}_b = [\mathbf{e}_{b,1}, \ldots, \mathbf{e}_{b,M}]^\top \in \mathbb{R}^{M \times d_{\text{vis}}}$, are aligned with their corresponding positive regions using an InfoNCE loss~\cite{oord2018representation}. For each \texttt{<SEG>} token, the paired region embedding $\mathbf{z}^{+}_{b,m}$ serves as a positive example, while image tokens from other images (and non-attended regions) act as negatives. With $L_2$-normalized vectors $\hat{\mathbf{e}}_{b,m}$ and $\hat{\mathbf{z}}^{+}_{b,m}$, the positive logit is $s^{+}_{b,m} = \langle \hat{\mathbf{e}}_{b,m},\, \hat{\mathbf{z}}^{+}_{b,m} \rangle$. The loss is defined as
\begin{equation}
\mathcal{L}_{\mathrm{NCE}}
= -\frac{1}{BM}\sum_{b,m}
\log \frac{\exp(a_{b,m})}{\exp(a_{b,m}) + \sum_{k \in \mathcal{K}^{-}} \exp(r_{b,mk})},
\end{equation}
where $a_{b,m} = s^{+}_{b,m}/\tau$, $r_{b,mk} = \langle \hat{\mathbf{e}}_{b,m},\, \hat{\mathbf{z}}^{-}_{k}\rangle / \tau$, and $\tau$ is the temperature parameter.


\subsection{PAVE: The VQA Dataset}

\noindent\textbf{Dataset Summary.}
We introduce \textbf{PAVE} (\textbf{P}edestrian \textbf{A}ccessibility and \textbf{V}isual-grounded \textbf{E}valuation), a spatially grounded VQA dataset for accessibility reasoning in complex pedestrian environments. It captures diverse real-world scenes with occlusions, reflections, motion blur, and dense object layouts typical of urban navigation. PAVE is built from the real-image subset of SANPO~\cite{waghmare2025sanpo}, which provides head-mounted pedestrian-view frames with human-annotated semantic and instance masks and corresponding depth maps. We focus exclusively on real images to preserve natural visual artifacts present in pedestrian scenes and therefore exclude synthetic data~\cite{waghmare2025sanpo,ying2025mmwalk}. SANPO’s panoptic labels are converted into unified semantic maps, while depth information is used separately to compute object-level distances for each feature. The dataset contains 41k image–question–answer triplets, each consisting of an RGB frame, a question about path accessibility, and a free-form answer describing accessible features, harmful features, their distances from the camera, and an overall accessibility assessment. Additional details are provided in the Appendix.

\begin{figure}[t]
    \centering
    \includegraphics[width=0.95\columnwidth]{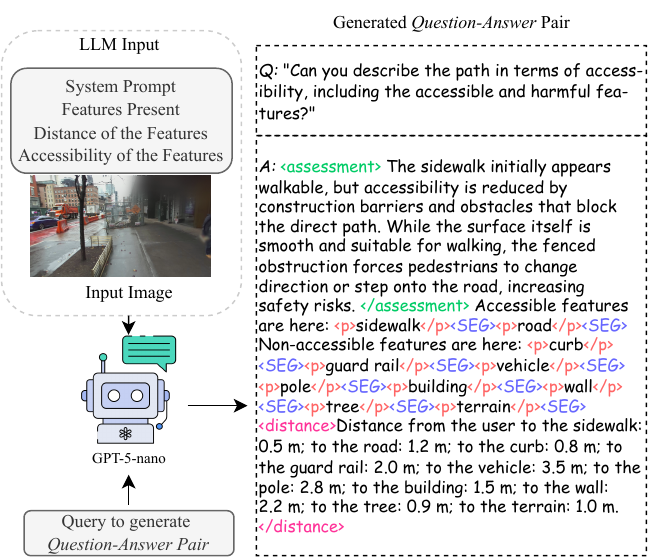}
    \caption{Pipeline for generating accessibility-aware VQA pairs in the PAVE dataset. The LLM receives the system prompt, detected features, their distance values, and the accessibility of the features, and generates structured outputs containing \textcolor{ForestGreen}{\texttt{<assessment>}}, 
    \textcolor{WalkPink}{\texttt{<distance>}}, \textcolor{blue}{\texttt{<SEG>}}, and
    \textcolor{red}{\texttt{<p>}} tokens.}
    \vspace{-5mm}
    \label{fig:dataset_curation}
\end{figure}

\noindent\textbf{Dataset Curation Pipeline.}
Each question–answer pair in PAVE is generated from a SANPO frame containing (1) the RGB image, (2) segmentation masks identifying visible objects/features, (3) depth information for each feature, and (4) accessibility labels (e.g., \textit{sidewalk} as accessible, \textit{vehicle} or \textit{stair} as harmful). For each feature mask, we compute the minimum pixel depth to represent its closest visible distance from the pedestrian viewpoint, providing object-level distance information for the final annotations. As shown in \Cref{fig:dataset_curation}, these structured scene attributes are then provided to GPT-5-nano~\cite{openai2025gpt5} to generate accessibility-aware VQA pairs. A system prompt instructs the model to act as a navigation assistant, with a one-shot example guiding the output format. GPT-5-nano produces both the natural-language question and a structured answer containing four elements: a concise \texttt{<assessment>} of overall accessibility, a \texttt{<distance>} tag listing object-level distances in text form, and \texttt{<p>} and \texttt{<SEG>} tokens that spatially ground the conversation. Additional details on prompt design and formatting are provided in the Appendix.

\noindent\textbf{Verification Pipeline.}
To ensure annotation reliability at scale, we combine automated validation with selective human review. Programmatic checks verify object references, \texttt{<SEG>} token associations, and consistency with sensor-derived depth across all samples. Manual inspection is then applied to the experimental subset to confirm structural correctness, including formatting and token usage.

\begin{figure*}[t]
    \centering
    \includegraphics[width=1.0\linewidth]{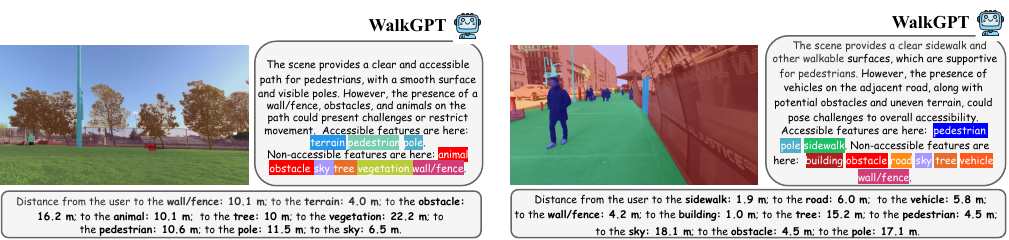}
    \caption{
    Qualitative results of WalkGPT on the PAVE validation set. Given a scene image, WalkGPT generates grounded conversations together with segmentation masks and depth-aware distance estimates, reflecting its understanding of accessibility and spatial context. Additional examples are provided in the Appendix.
    }

    \label{fig:qual_figure}
\end{figure*}

\subsection{The Training Recipe}

\noindent\textbf{Conversation Generation.}
The causal language model is trained autoregressively with teacher forcing, predicting each output token conditioned on preceding tokens. Let $x$ denote the concatenated textual input (system prompt, question, and prior outputs), and let $\mathbf{V}_{\text{proj}}$ denote the projected image tokens produced by the MSQP from the frozen pixel encoder (see Section~\ref{sec:methods}). The model generates the answer sequence $y_{1:S}$ by jointly attending to textual and visual embeddings through cross-attention. The training objective is the standard cross-entropy loss,
\begin{equation} 
\mathcal{L}_{\mathrm{CE}}
= -\frac{1}{S}\sum_{s=1}^{S} \log P\!\left(y_s \mid y_{<s},\, x,\, \mathbf{V}_{\text{proj}}\right),
\end{equation}
where $S$ denotes the number of answer tokens. Tokens from the prompt and question contribute to the conditioning context but are excluded from loss computation. Structured tokens provide task-specific control over the generated output: \texttt{<assessment>} produces an accessibility judgment, \texttt{<p>} grounds referenced objects within the dialogue, \texttt{<SEG>} provides the segmentation prompt for the pixel decoder, and \texttt{<distance>} expresses object-level depth in text form. Representing these outputs as language tokens allows WalkGPT to unify conversational reasoning, spatial grounding, segmentation prompting, and depth description within a single next-token prediction process.

\noindent\textbf{Segmentation Mask Prediction.}
As illustrated in \Cref{fig:architecture}a, \texttt{<SEG>} tokens generated within the textual response serve as spatial prompts for the pixel decoder, enabling segmentation grounded in the language output. Each \texttt{<SEG>} token corresponds to a semantic entity mentioned in the answer and is associated with an image region. The hidden state of each \texttt{<SEG>} token is mapped by the CTP into the visual space and used by the decoder together with the projected visual embeddings $\mathbf{V}_{\text{proj}}$ to predict one mask per token, following~\cite{rasheed2024glamm}. Segmentation supervision combines Dice and cross-entropy losses,
$\mathcal{L}_{\text{seg}} = \mathcal{L}_{\text{Dice}} + \mathcal{L}_{\text{CE}_{seg}}$,
encouraging accurate mask prediction and alignment between linguistic entities and visual regions.

\noindent\textbf{Depth Estimation.}
WalkGPT predicts object-level depth as part of the same autoregressive sequence used for grounded conversation. Distance information is expressed through the \texttt{<distance>} \dots \texttt{</distance>} span of the generated answer, where objects referenced in the response are associated with textual distance descriptions. During training, sensor-derived depth maps are used to compute the minimum visible distance for each segmented object, which is inserted into the ground-truth answers as discretized natural-language distance expressions (\Cref{fig:dataset_curation}). These tokens are learned through the same cross-entropy objective used for conversation generation. Because the referenced objects are already grounded through \texttt{<SEG>} tokens, the model learns to associate segmentation-aligned visual regions with their corresponding distance descriptions. Repeated supervision of these distance statements encourages attention to visual cues correlated with relative depth, allowing depth estimation to emerge through language prediction without dense depth supervision or dedicated depth heads.

\noindent\textbf{Overall Training Objective.}
The final objective integrates three complementary components: (1) cross-entropy loss $\mathcal{L}_{\mathrm{CE}}$ for grounded conversation generation, (2) segmentation loss $\mathcal{L}_{\mathrm{seg}}$ for mask prediction, and (3) contrastive alignment loss $\mathcal{L}_{\mathrm{NCE}}$ for visual–textual correspondence. The total loss is expressed as
\begin{equation} 
\mathcal{L}_{\text{total}} = 
\alpha_1\,\mathcal{L}_{\mathrm{CE}} +
\alpha_2\,\mathcal{L}_{\mathrm{seg}} +
\alpha_3\,\mathcal{L}_{\mathrm{NCE}},
\end{equation} 

\noindent where $\alpha_1$, $\alpha_2$, and $\alpha_3$ are scalar weights controlling each component’s contribution to jointly optimize conversation generation, spatial grounding, and cross-modal alignment.

%% file: sec/4_experiments.tex
\section{Experiments}
\label{sec:experiments}
\subsection{Implementation Details}

WalkGPT uses a single SAM ViT-H pixel encoder~\cite{kirillov2023segment} shared across all components, providing consistent visual grounding for both text generation and mask prediction. The language model is initialized with pretrained checkpoints from~\cite{ren2024pixellm} for the 13B version and~\cite{rasheed2024glamm} for the 7B version. We apply LoRA~\cite{ding2023} to fine-tune the language model parameters in a lightweight and efficient manner. All experiments are run on a single NVIDIA H100 GPU (80 GB) with Python 3.10.8 and CUDA 12.1. The PAVE dataset contains 91 pedestrian-view video sessions recorded at high frame rates. Because frames are sequential, consecutive images often change very little, which can create strong temporal redundancy that can encourage memorization during training. To obtain a more balanced representation of each scene, we uniformly sample 100 frames per session. This reduces redundancy while preserving scene-level diversity. We use 85 sessions for training (about 8.5k frames) and hold out 6 sessions (around 600 frames) for evaluation.

Training proceeds in two stages. In the pretraining stage, only the MSQP module is optimized while all other components remain frozen, allowing MSQP to learn stable visual tokenization across heterogeneous datasets. We use ADE20K~\cite{zhou2019semantic} and the RefCOCO family~\cite{kazemzadeh2014referitgame} for this stage. The fine-tuning stage jointly trains MSQP, CTP, the pixel decoder, and the LoRA parameters on the PAVE dataset. All models share the same optimization settings for fair comparison. Additional hyperparameter details and ablations are provided in the Appendix.


\begin{table*}[t]
  \centering
  \small
  \setlength{\tabcolsep}{3pt}
  \renewcommand{\arraystretch}{1.1}
  \caption{Performance comparison on grounded navigation conversation generation. 
Models marked with \textsuperscript{\dag} are zero-shot; “-FT” indicates fine-tuned on PAVE. 
Depth metrics for zero-shot models are listed as N/A because they fail to produce any depth estimations. Best results are \textbf{bold-faced.}}

  \label{tab:main_results}
  \resizebox{1.5\columnwidth}{!}{
  \begin{tabular}{l|cc|ccc|cc}
    \hline
    \multirow{2}{*}{Model} &
    \multicolumn{2}{c|}{Text Generation} &
    \multicolumn{3}{c|}{Segmentation Performance} &
    \multicolumn{2}{c}{Depth Estimation} \\
    \cline{2-3} \cline{4-6} \cline{7-8}
     & CIDEr$\uparrow$ & METEOR$\uparrow$ & AP50$\uparrow$ & mIoU$\uparrow$ & Recall$\uparrow$ & Depth Acc.$\uparrow$& AbsRel$\downarrow$ \\
    \hline
    \hline
    GLAMM\textsuperscript{\dag}~\cite{rasheed2024glamm}        & 1.32 & 21.98 & 1.23  & 2.01 & 2.24 & N/A& N/A\\
    GLAMM-FT~\cite{rasheed2024glamm}                           & 37.96 & 39.12 & 15.21 & 18.23 & 25.01 &38.95 & 77.05\\
    LISA\textsuperscript{\dag}~\cite{lai2024lisa}              & 0.97 &  17.75 & 1.02 &  1.50 & 1.84 &N/A&N/A\\
    LISA-FT~\cite{lai2024lisa}                                 & 35.14 & 36.17 & 13.71 & 15.07 & 24.11 &35.46& 81.22 \\
    PixelLM\textsuperscript{\dag}~\cite{ren2024pixellm}        & 1.08 &21.87 & 1.22 & 1.59 & 2.39 &N/A &N/A\\
    PixelLM-FT~\cite{ren2024pixellm}                           & 37.49 & 38.02 & 15.97 & 18.10 & 28.92 & 39.00 &74.61\\
    GSVA\textsuperscript{\dag}~\cite{xia2024gsva}              & 0.78 & 20.74 & 1.78 & 1.87 & 2.54 &N/A &N/A\\
    GSVA-FT~\cite{xia2024gsva}                                 & 35.78 & 38.15 & 14.67 & 17.34 & 29.71 & 36.55& 78.19\\
    OMG-LLaVA\textsuperscript{\dag}~\cite{zhang2024omg}        & 0.97 & 19.99& 2.05 & 3.21 & 2.02 &N/A &N/A \\
    OMG-LLaVA-FT~\cite{zhang2024omg}                           & 38.01 & 38.96& 15.74 & 18.02 & 28.05 & 39.02 & 75.01 \\
    Sa2VA\textsuperscript{\dag}~\cite{yuan2025sa2va}        & 1.28 & 21.02& 2.71 & 3.14 & 1.78 &N/A &N/A \\
    Sa2VA-FT~\cite{yuan2025sa2va}                           & 38.82 & 39.66& 18.72 & 16.10 & 29.20 & 40.54 & 73.82 \\
    \hline
    \rowcolor{gray!15}
    WalkGPT (7B)                                      & \textbf{41.97} & 42.36 & 16.66 & 19.95 & 31.55 & 41.97 & \textbf{67.88}\\
    \rowcolor{gray!15}
    WalkGPT (13B)                                     & 41.17 & \textbf{43.01} & \textbf{17.26} & \textbf{20.16} & \textbf{32.71} & \textbf{48.95}&70.66 \\
    \hline
  \end{tabular}
  }
\end{table*}

\subsection{Baselines}
\label{sec:baselines}

We compare WalkGPT with leading grounded vision-language models, including GLAMM~\cite{rasheed2024glamm}, LISA~\cite{lai2024lisa}, PixelLM~\cite{ren2024pixellm}, GSVA~\cite{xia2024gsva}, OMG-LLaVA~\cite{zhang2024omg}, and Sa2VA~\cite{yuan2025sa2va}, using the validation split of the PAVE dataset. Each model is evaluated in both zero-shot and fine-tuned settings and adapted to our task by extending its tokenizer with our four structured token types (\texttt{<SEG>}, \texttt{<distance>}, \texttt{<p>}, \texttt{<assessment>}), ensuring all baselines operate under the same multimodal interface. We evaluate three aspects of grounded navigation: text generation, segmentation, and depth estimation. CIDEr and METEOR measure text quality, while AP50, mIoU, and Recall assess segmentation of objects of interest. Depth Accuracy and Absolute Relative Error (AbsRel) quantify depth estimation. Depth Accuracy measures the proportion of estimations within $[0.5\times, 2\times]$ of ground truth, and AbsRel computes the mean absolute difference normalized by ground-truth depth. Additional metric details are provided in the Appendix. To assess generalization beyond navigation, we also evaluate Referring Expression Segmentation (RES) on RefCOCO, RefCOCO+, and RefCOCOg datasets. Following standard protocols~\cite{lai2024lisa,ren2024pixellm}, we report precision at multiple IoU thresholds and mean IoU, and compare against baselines including MCN~\cite{luo2020multi}, VLT~\cite{ding2021vision}, CRIS~\cite{wang2022cris}, LAVT~\cite{yang2022lavt}, ReLA~\cite{liu2023gres}, LISA~\cite{lai2024lisa}, and PixelLM~\cite{ren2024pixellm}.



\subsection{Results}

\begin{table}[t]
  \centering
  \small
  \setlength{\tabcolsep}{6pt}
  \renewcommand{\arraystretch}{1.15}
  \caption{Performance comparison on the referring expression segmentation (RES) benchmark. Best results are \textbf{bold-faced.}}
  \label{tab:refseg}
  \resizebox{\columnwidth}{!}{
  \begin{tabular}{l ccc ccc cc}
    \toprule
    \multirow{2}{*}{\textbf{Method}} 
      & \multicolumn{3}{c}{\textbf{refCOCO}} 
      & \multicolumn{3}{c}{\textbf{refCOCO+}} 
      & \multicolumn{2}{c}{\textbf{refCOCOg}} \\
    \cmidrule(lr){2-4} \cmidrule(lr){5-7} \cmidrule(lr){8-9}
      & \textbf{val} & \textbf{testA} & \textbf{testB}
      & \textbf{val} & \textbf{testA} & \textbf{testB}
      & \textbf{val(U)} & \textbf{test(U)} \\
    \midrule
    MCN~\cite{luo2020multi}        & 62.4 & 64.2 & 59.7 & 50.6 & 55.0 & 44.7 & 49.2 & 49.4 \\
    VLT~\cite{ding2021vision}       & 67.5 & 70.5 & 65.2 & 56.3 & 61.0 & 50.1 & 55.0 & 57.7 \\
    CRIS~\cite{wang2022cris}      & 70.5 & 73.2 & 66.1 & 62.3 & 68.1 & 53.7 & 59.9 & 60.4 \\
    LAVT~\cite{yang2022lavt}      & 72.7 & 75.8 & 68.8 & 62.8 & 68.4 & 55.1 & 61.2 & 62.1 \\
    ReLA~\cite{liu2023gres}      & 73.8 & 76.5 & 70.2 & 66.0 & 71.0 & 57.7 & 65.5 & 66.0 \\
    \hline
    \rowcolor{gray!15}
    LISA~\cite{lai2024lisa}       & 74.1 & 76.5 & 71.1 & 62.4 & 67.4 & 56.5 & 66.4 & 68.5 \\
    \rowcolor{gray!15}
    PixelLM~\cite{ren2024pixellm}    & 73.0 & 76.5 & 68.2 & 66.3 & \textbf{71.7} & 58.3 & 69.3 & 70.5 \\
    \rowcolor{gray!15}
    \textbf{WalkGPT (Ours)} &  \textbf{76.2}  &  \textbf{78.5}  &  \textbf{68.3}  &  \textbf{70.0}  &  71.1  &  \textbf{60.5}  &  \textbf{72.6}  &  \textbf{71.6}  \\
    \hline
  \end{tabular}
  }
  \vspace{-5mm}
\end{table}


\noindent\textbf{Grounded Navigation Conversation Generation.}
\Cref{fig:qual_figure} shows qualitative examples of WalkGPT generating grounded navigation conversations from single pedestrian-view images. The model provides an overall assessment and segments \textit{accessible} and \textit{harmful} features, including their distances from the user, helping pedestrians understand what to avoid and how far potential obstacles are. These examples span visually distinct scenes and illustrate that WalkGPT maintains consistent spatial grounding across different layouts. \Cref{tab:main_results} summarizes performance on text generation, segmentation, and depth estimation. Zero-shot baselines fail across all metrics and cannot predict depth even with added \texttt{<distance>} tokens, highlighting the difficulty of this real-world task. Fine-tuning improves results but remains inadequate. WalkGPT achieves substantial gains over strong fine-tuned baselines such as PixelLM-FT and OMG-LLaVA-FT: the 13B model improves mIoU by more than 10\% (20.16 vs. 18.10) and raises depth accuracy by over 25\% (48.95 vs. 39.00), while the 7B variant surpasses prior best text-generation scores by over 10\% in CIDEr. These improvements stem from WalkGPT’s unified pixel encoder together with MSQP and CTP, where MSQP provides fine-grained multi-scale visual tokens and CTP, guided by region alignment, maintains consistent grounding between visual regions and structured outputs. Together, they support robust multimodal grounding across model sizes.

\noindent\textbf{Referring Expression Segmentation (RES).}
RES requires segmenting target regions in an image given a natural-language referring expression. Standard RES models~\cite{luo2020multi,ding2021vision,wang2022cris,yang2022lavt,liu2023gres,lai2024lisa,ren2024pixellm,zou2023generalized,zou2023segment} take the expression directly as input. For WalkGPT and the grounded LVLMs in \Cref{tab:refseg}, we place the expression into a short instruction template and prompt the model to generate a response containing a \texttt{<SEG>} token, whose embedding is decoded by the pixel decoder to produce the mask. Following the setup in \Cref{sec:baselines}, we evaluate WalkGPT on public RES benchmarks. Although not designed specifically for RES, WalkGPT shows strong generalization as a grounded vision–language model. As shown in \Cref{tab:refseg}, it achieves 76.2\% on refCOCO-val and 72.6\% on refCOCOg-val, outperforming LISA and PixelLM by up to about 3--4\%. These gains reflect stronger visual–language grounding, enabling more precise localization of the referred region without RES-specific training.



\begin{table}[H]
\centering
\small
\setlength{\tabcolsep}{8pt}
\vspace{-3mm}
\caption{Segmentation performance on PAVE compared with representative vision-only segmentation benchmarks.}
\resizebox{0.7\columnwidth}{!}{
\begin{tabular}{lcc}
\toprule
Model & mIoU$\uparrow$ & Recall$\uparrow$\\
\midrule
U-Net~\cite{ronneberger2015unet} & 16.85 & 28.34 \\
nnU-Net~\cite{isensee2021nnu} & 18.55 & 28.41 \\
Swin-UNETR~\cite{hatamizadeh2021swin} & \textbf{20.60} & 30.65 \\
\hline
\rowcolor{gray!15}
WalkGPT (Ours) & 20.16 & \textbf{32.71} \\
\hline
\end{tabular}}
\vspace{-3mm}
\label{tab:vision_only_comparison}
\end{table}

\begin{figure}[H]
    \centering
    \includegraphics[width=\columnwidth]{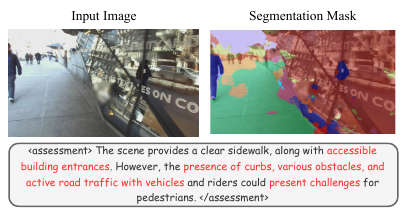}
    \caption{Failure case study on PAVE. WalkGPT misinterprets strong road reflections on the building façade as physical obstacles, producing \textcolor{red}{incorrect guidance} even though the path itself is fully accessible. Part of the image is blurred for privacy.}
    \label{fig:failure}
\end{figure}

\noindent\textbf{Challenges and Failure Analysis.}
To contextualize segmentation performance on PAVE, we compare WalkGPT with strong vision-only benchmarks, including U-Net~\cite{ronneberger2015unet}, nnU-Net~\cite{isensee2021nnu}, and Swin-UNETR~\cite{hatamizadeh2021swin}. Although specialized for segmentation, these models achieve under 21\% mIoU, reflecting the difficulty of PAVE’s dense pedestrian-view scenes characterized by heavy occlusions, small regions of interest, and severe class imbalance. WalkGPT attains a comparable mIoU (20.16) without segmentation-specific fine-tuning, indicating that even expert models struggle under these conditions. These results underscore the inherent difficulty of PAVE’s real-world scenes for grounded segmentation in navigation settings. \Cref{fig:failure} provides a representative example: strong road reflections on the building façade resemble physical obstacles, causing WalkGPT to segment them as real objects and generate flawed accessibility guidance. This misinterpretation propagates to the associated segmentation output, further degrading mask quality. Such failures are particularly common in single-view images, where reflective surfaces distort visual depth cues and make it difficult to distinguish true geometry from appearance. Similar issues arise from motion blur, noisy surfaces, and significant class imbalance (see Appendix), all of which blur the boundary between true structures and incidental visual artifacts.


\begin{table}[b]
\centering
\small
\setlength{\tabcolsep}{8pt}
\vspace{-3mm}
\caption{Comparison of hallucination ($CHAIR_i$) and object coverage (Cover) scores across LVLMs on the PAVE dataset.}
\resizebox{0.7\columnwidth}{!}{\begin{tabular}{lcc}
\toprule
Model & $CHAIR_i$$\downarrow$ & Cover$\uparrow$ \\
\midrule
LLaVA-1.5~\cite{liu2024improved} & 22.16 & 33.04 \\
LLaVA 1.6 Mistral~\cite{li2024llava} & 23.56 & 38.83 \\
Qwen-VL-Chat~\cite{bai2023qwen} & 26.78 & 31.42 \\
\hline
\rowcolor{gray!15}
WalkGPT (Ours) & \textbf{18.49} & \textbf{83.66}  \\
\hline
\end{tabular}}

\label{tab:hallucination_comparison}
\end{table}

\noindent\textbf{Grounded LVLM for Hallucination Mitigation.}
\Cref{tab:hallucination_comparison} reports hallucination performance against non-grounded LVLMs. LLaVA-1.5~\cite{liu2024improved}, LLaVA~1.6~Mistral~\cite{li2024llava}, and Qwen-VL-Chat~\cite{bai2023qwen} show high hallucination rates and limited coverage of visible objects, reflecting their lack of explicit grounding mechanisms. WalkGPT incorporates pixel-level grounding by linking text generation to segmentation-informed visual features. This connection reduces unsupported object mentions and improves recognition of image regions that are actually present, leading to substantially lower $CHAIR_i$ and higher Cover scores. Constraining language to visual evidence allows WalkGPT to produce scene descriptions that remain faithful to the underlying image.

\begin{table}[t]
\centering
\small
\setlength{\tabcolsep}{8pt}
\caption{Ablation study examining the impact of different design choices in WalkGPT.}
\resizebox{1.0\columnwidth}{!}{\begin{tabular}{lccc}
\toprule
Variant & METEOR$\uparrow$ & mIoU$\uparrow$ &Depth Acc.$\uparrow$\\
\midrule
WalkGPT (Full) & \textbf{43.01} & \textbf{20.16} &\textbf{48.95}\\
\quad w/o MSQP $\rightarrow$ MLP & 39.50 & 17.40 &43.39\\
\quad w/o MSQP multi-scale & 41.60 & 19.30 &44.70\\
\quad MSQP queries $Q{=}8$ & 38.10 & 16.20 &45.33\\
\quad CTP $\rightarrow$ Linear & 40.70 & 18.60 &47.98\\
\quad w/o $L_{\text{NCE}}$ & 41.00 & 18.90 &47.00\\
\quad w/o LoRA (LLM frozen) & 35.20 & 17.80 &40.21\\
\quad w/o \texttt{<distance>} & 41.22 & 20.01 &38.77\\
\bottomrule
\end{tabular}}
\vspace{-5mm}
\label{tab:ablate_components}
\end{table}

\noindent\textbf{Ablation Study.}
We evaluate several design variations of WalkGPT to assess their impact on grounded navigation. \Cref{tab:ablate_components} shows that replacing MSQP with a simple MLP leads to a clear drop in performance (43.01→39.50 METEOR, 20.16→17.40 mIoU, 48.95→43.39 Depth Acc.), highlighting MSQP as a key component. Removing its multi-scale aggregation also reduces performance (41.60 METEOR, 19.30 mIoU, 44.70 Depth Acc.), confirming that multi-resolution features are important for capturing spatial structure and depth cues. Reducing the number of MSQP queries to $Q{=}8$ (from $Q{=}32$) yields a further decline (38.10 METEOR, 16.20 mIoU, 45.33 Depth Acc.), indicating that query diversity supports robust grounding and distance reasoning. Substituting CTP with a linear mapping (40.70 METEOR, 18.60 mIoU, 47.98 Depth Acc.) or removing the contrastive loss $L_{\text{NCE}}$ (41.00 METEOR, 18.90 mIoU, 47.00 Depth Acc.) mainly affects language and segmentation while leaving depth largely unchanged. Freezing the LLM reduces performance across metrics (35.20 METEOR, 17.80 mIoU, 40.21 Depth Acc.), suggesting that lightweight language adaptation also supports distance-aware guidance. Finally, removing the \texttt{<distance>} token primarily harms depth prediction (48.95→38.77) while leaving segmentation nearly unchanged (20.16→20.01 mIoU), confirming the importance of explicit distance supervision. Overall, these results confirm that WalkGPT’s performance arises from the joint contribution of MSQP, CTP-based alignment, and structured distance tokens.

%% file: sec/5_conclusion.tex
\section{Conclusion}
\label{sec:conclusion}
WalkGPT advances grounded multimodal reasoning by reframing pedestrian navigation as an interpretable, accessibility-aware dialogue grounded in pixels and depth. By jointly modeling conversational guidance, segmentation, and distance estimation, and introducing the PAVE dataset, it establishes a benchmark for accessibility-aware reasoning and opens new directions for trustworthy assistive navigation systems. Beyond this task, our framework highlights the importance of tightly coupling language reasoning with spatial grounding for real-world multimodal AI.

\noindent\textbf{Limitations and Future Work.}
WalkGPT may still be affected by dataset artifacts that introduce ambiguity. Future work will explore improved depth estimation and evaluate cross-domain generalization on additional navigation and grounding datasets.

\section*{Acknowledgments}
This work was supported by the National Institutes of Health (NIH), National Eye Institute (NEI), under Grant R61EY037504. 


%% file: sec/X_suppl.tex
\clearpage
\setcounter{page}{1}
\maketitlesupplementary

\section{Implementation Details}
\subsection{Hyperparameter Settings.}

\noindent\textbf{Training configuration.}
WalkGPT is trained for 10 epochs with a batch size of 16 and a gradient accumulation factor of 10, resulting in an effective batch size of 160 samples. The optimizer is AdamW with a learning rate of $2\times10^{-4}$. All experiments use \texttt{bf16} precision and a maximum sequence length of 2048 tokens. Images are resized to a resolution of $448\times448$ before being processed by the vision encoder. Each epoch consists of 54 optimization steps, corresponding to the SANPO training split used in our setup.

\noindent\textbf{Segmentation and alignment objectives.}
The segmentation branch optimizes a combination of Dice and BCE losses over the predicted masks. In addition, WalkGPT employs a contrastive alignment objective that pairs text-side \texttt{<SEG>} token embeddings with local SAM features. SAM produces 256-dimensional visual tokens, which are flattened and projected into the LLM hidden space using the Multi-Scale Query Projector (MSQP), configured with a $6\times6$ target token grid.

\noindent\textbf{Loss weighting and contrastive settings.}
The overall objective follows the formulation described in the main paper, with loss weights $\alpha_1=0.1$ for the CE loss, $\alpha_2=0.05$ and $0.35$ for the Dice and BCE segmentation losses respectively, and $\alpha_3=0.3$ for the InfoNCE alignment term. The InfoNCE loss uses a temperature of $0.07$ and top-$8$ hard negative selection when computing contrastive similarities.

\noindent\textbf{Query and projection modules.}
MSQP operates in a 1024-dimensional hidden space and uses two cross-attention layers per scale (8 attention heads), with a total of 32 queries allocated as 12/8/8/4 across $1\times$, $2\times$, $4\times$, and global scales. The resulting tokens are padded to a $6\times6$ grid before projection into the LLM embedding space. CTP is implemented as a calibrated MLP projector with widen factor 2 and LayerNorm, and applies a learned temperature (logit scale) to normalized text embeddings.

\subsection{Computational Statistics.}
We report the computational characteristics of WalkGPT to provide transparency regarding training and inference costs. All statistics correspond to the final configuration used in our experiments and are not intended as comparative benchmarks. The model contains approximately 14.1B parameters in total. Training was performed for 10 epochs on the 8.5k-sample SANPO training split, requiring approximately 6 hours on 8 GPUs. Inference throughput was measured independently, with 1k queries processed in approximately 1 hour under the same hardware configuration.

\subsection{Structured Token Design}
We introduce four categories of structured tokens to extend the language model vocabulary and enable multimodal grounding and spatial reasoning for the navigation task.

\begin{itemize}[leftmargin=10pt]

    \item \textbf{\texttt{<assessment>} and \texttt{</assessment>} Tokens:}
    These tags enclose a concise qualitative summary of scene accessibility, encouraging the model to generate natural language evaluations of how walkable or obstructed the environment appears.

    \item \textbf{\texttt{<SEG>} Tokens:}
    These tokens indicate objects referenced in the response that correspond to pixel-level segmentation regions. During training, each \texttt{<SEG>} token is aligned with its ground-truth mask to provide spatial grounding and interpretable visual--text associations.

    \item \textbf{\texttt{<p>} and \texttt{</p>} Tokens:}
    These tags wrap short descriptive phrases associated with specific visual elements, enabling phrase-level grounding by linking textual mentions to the corresponding regions in the image.

    \item \textbf{\texttt{<distance>} and \texttt{</distance>} Tokens:}
    These tags encode relative distances derived from SANPO depth maps, allowing the model to associate textual references with spatial proximity and improving depth-aware reasoning.

\end{itemize}

\begin{figure*}[t]
    \centering
    \includegraphics[width=1.0\linewidth]{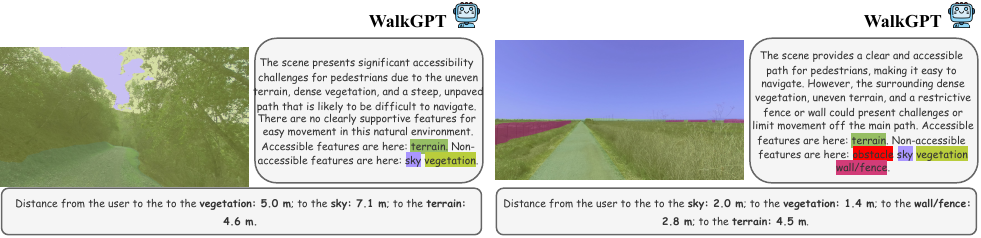}
    \caption{
    Additional qualitative results of WalkGPT on the PAVE validation set for off-road scenes. Examples illustrate the model’s ability to handle unstructured outdoor environments with uneven terrain,
    dense vegetation, and limited walkable surfaces.
    }

    \label{fig:qual_figure_2}
    \vspace{-5mm}
\end{figure*}

\subsection{Depth Estimation Metrics.}
To evaluate the numerical depth predictions produced during conversation, we introduce two complementary metrics: \textbf{Depth Accuracy (Depth Acc.)} and \textbf{Absolute Relative Error (AbsRel)}. Let $d_i^{\text{pred}}$ and $d_i^{\text{gt}}$ denote the predicted and ground-truth depths for object $i$, respectively, and let $N$ be the total number of evaluated objects.

Depth Acc. measures the proportion of predictions that fall within a multiplicative tolerance of the ground-truth depth. Specifically, a prediction is considered correct if
\begin{equation}
0.5 \times d_i^{\text{gt}} \;\le\; d_i^{\text{pred}} \;\le\; 2 \times d_i^{\text{gt}}.
\end{equation}
The metric is computed as
\begin{equation}
\text{Depth Acc.} = \frac{1}{N} \sum_{i=1}^{N} \mathbf{1}\!\left( 0.5\, d_i^{\text{gt}} \le d_i^{\text{pred}} \le 2\, d_i^{\text{gt}} \right),
\end{equation}
where $\mathbf{1}(\cdot)$ denotes the indicator function.

Absolute Relative Error (AbsRel) provides a scale-normalized measure of depth discrepancy and is defined as
\begin{equation}
\text{AbsRel} = \frac{1}{N} \sum_{i=1}^{N} \frac{\left| d_i^{\text{pred}} - d_i^{\text{gt}} \right|}{d_i^{\text{gt}}}.
\end{equation}
Together, Depth Acc. captures coarse correctness within a reasonable interval, while AbsRel measures the relative magnitude of depth error with respect to the ground-truth value.

\begin{figure}[H]
    \centering
    \includegraphics[width=\columnwidth]{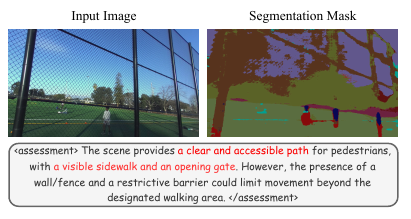}
    \caption{Another failure case on PAVE. WalkGPT incorrectly infers that the fenced area provides an open and accessible path, misled by the transparency of the fence and the clear view of the space behind it.}
    \label{fig:failure_appendix}
\end{figure}

\section{Additional Qualitative Results}
\Cref{fig:qual_figure_2} presents additional qualitative examples from the PAVE dataset, highlighting diverse off-road scenes and their corresponding accessibility annotations. \Cref{fig:failure_appendix} shows a representative failure case where WalkGPT misinterprets a fenced boundary as an open, walkable passage due to the fence’s transparency.

\section{Dataset Details}
\subsection{PAVE Dataset}

\noindent\textbf{SANPO: Summary.}
The source imagery dataset, SANPO \cite{waghmare2025sanpo}, provides large-scale egocentric video captured from eye-level and chest-level viewpoints using stereo cameras mounted on real volunteer runners. Each session contains synchronized left--right video streams, associated camera poses, and both sparse depth (from the ZED sensor) and dense depth estimated with CREstereo. SANPO also includes temporally consistent panoptic segmentation for a subset of frames, high-level session attributes (e.g., environment type, visibility, motion), and hardware/IMU metadata. In addition to real captures, the dataset provides 113K synthetic frames generated under similar conditions, enabling controlled comparisons between real and simulated environments. All recordings follow strict privacy and legal guidelines, including participant review and automatic blurring of personally identifiable information.

\begin{figure}[H]
    \centering
    \includegraphics[width=\columnwidth]{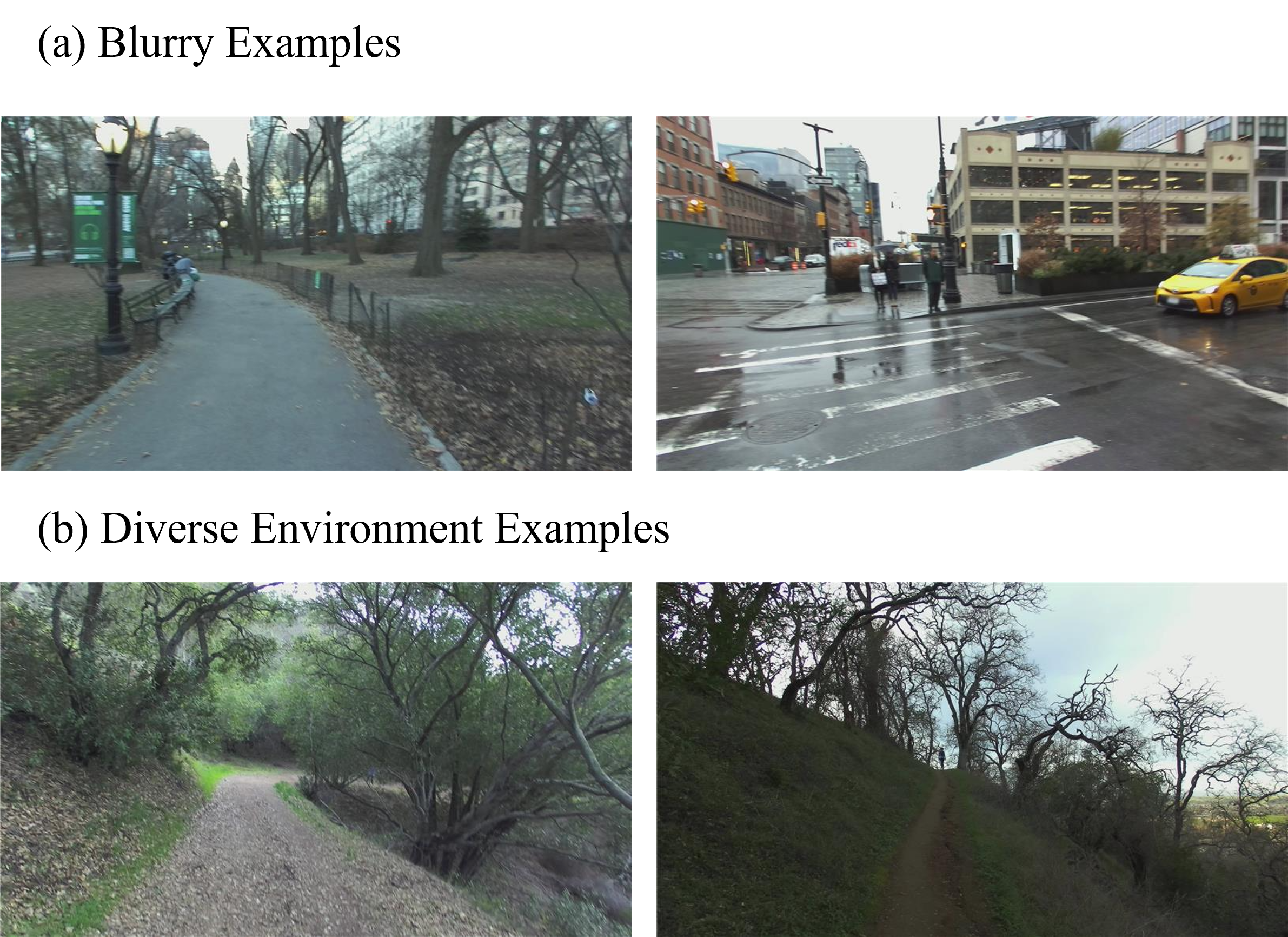}
    \caption{Qualitative examples illustrating varied capture conditions in SANPO. (a) Motion blur and imaging artifacts. (b) Diverse outdoor environments spanning urban streets, parks, and natural trails.}
    \label{fig:dataset_diversity}
\end{figure}

\noindent\textbf{SANPO: Geographic and environmental coverage.}
SANPO-Real consists of 701 real-world egocentric recording sessions collected across four geographically distinct locations in the United States: San Francisco (CA), Mountain View (CA), Boulder (CO), and New York City (NY). These regions were selected to capture a diverse mix of urban cores, suburban neighborhoods, public parks, road junctions, and open pedestrian spaces. Recordings span a wide range of environmental conditions, including sunny, cloudy, rainy, and snowy weather, as well as variations in visibility, elevation change (flat, uphill, downhill, stairs), ground appearance (e.g., asphalt, pavers, gravel, terrain), and pedestrian and vehicular traffic density. Sessions also vary in time of day and motion patterns, covering walking, jogging, and running with different levels of motion blur. Figure~\Cref{fig:dataset_diversity} illustrates representative scenes across urban streets, park pathways, and narrow dirt trails in vegetation-dense environments.

\begin{figure}[H]
    \centering
    \includegraphics[width=\columnwidth]{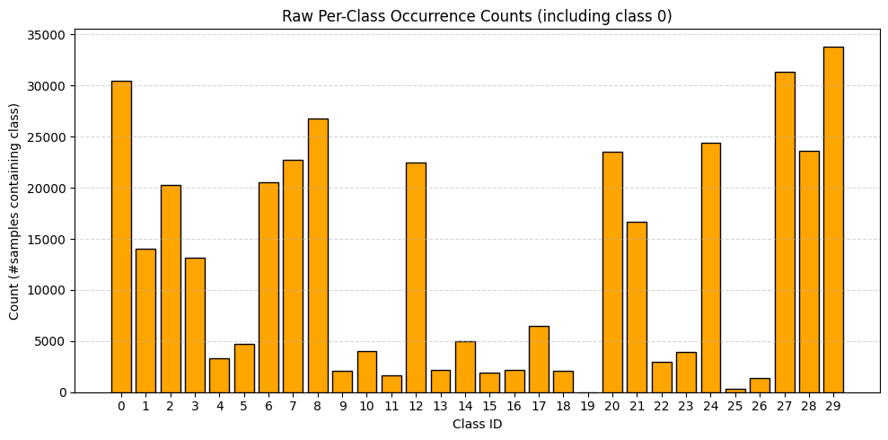}
    \caption{Per-class sample occurrence counts across all semantic categories (including background class 0). The x-axis denotes class IDs and the y-axis indicates the number of samples containing each class.}
    \label{fig:dataset_distribution}
\end{figure}

\noindent\textbf{Labels.}
SANPO defines 30 categories spanning both semantic and panoptic annotation types, including \emph{road} (1, semantic), \emph{curb} (2, semantic), \emph{sidewalk} (3, semantic), \emph{crosswalk} (5, panoptic), \emph{building} (7, semantic), \emph{pedestrian} (12, panoptic), \emph{vehicle} (21, panoptic), \emph{tree} (28, panoptic), and additional walkability-relevant classes such as \emph{stairs} (15, panoptic), \emph{obstacle} (20, panoptic), and \emph{other walkable surface} (17, semantic). \Cref{fig:dataset_distribution} shows the per-class sample occurrence counts, highlighting strong class imbalance across labels. Many walkability-critical classes appear far less frequently than dominant background and surface categories, making dense segmentation particularly challenging.

\noindent\textbf{Processing.}
We use only SANPO-Real frames that include human-annotated masks, covering both semantic-only and panoptic-encoded categories. For classes annotated in panoptic format, we convert the 3-channel PNG masks into single-channel semantic masks by extracting the semantic ID from the first channel and ignoring instance identifiers. Semantic-only masks are retained as provided. When resizing is required, we apply nearest-neighbor interpolation to preserve label integrity and clamp values to the valid ID range $\{0,\dots,30\}$. This yields a unified semantic representation suited for accessibility reasoning, which relies on class-level occupancy rather than instance differentiation.

\noindent\textbf{Depth Estimation.}
For each SANPO-Real frame, we use the corresponding dense depth map to compute a per-class distance from the camera and store it as ground truth for dataset construction. Because each semantic region may contain many scattered pixels, we derive a single representative depth value by taking the minimum depth among all pixels belonging to that class. This choice emphasizes the closest visible surface of each object, which is most relevant for accessibility reasoning and near-field obstacle assessment.

\subsection{Prompt to Generate PAVE}
To construct consistent natural-language annotations for pedestrian accessibility, we employ a large language model (LLM) to generate both the user-facing question and the structured answer associated with each scene. The generation pipeline operates in two stages. In the first stage, the LLM receives the RGB image (encoded in base64 format) through the vision-enabled GPT-5-nano API, together with a system prompt and a single formatted example. The model generates (i) a natural conversational question a pedestrian might ask about the environment and (ii) a short answer whose first block is a qualitative \texttt{<assessment>} describing overall walkability based solely on visual cues. All internal metadata (class labels, IDs, and distances) are explicitly withheld from the LLM. The JSON output is automatically validated, and malformed responses trigger a regeneration attempt.

In the second stage, the automatically generated assessment is augmented using ground-truth semantic and geometric information. Each object present in the frame is assigned to either the supportive (accessible) or harmful (non-accessible) category according to a fixed label-to-type mapping defined by the PAVE ontology. Depth values are derived from SANPO-Real dense depth maps; for each object, a representative distance is computed as the minimum depth across its pixels, corresponding to the closest visible surface and reducing occlusion bias. These elements are inserted into a fixed template to produce the final question–answer pair.

\noindent\textbf{Prompt Specification for Accessibility Question-Answer Generation.}
The LLM is instructed to behave as a navigation assistant that generates a natural question and a structured answer in a predefined format. The question must reference helpful and harmful scene elements in general terms, remain user-facing and conversational, and avoid any internal metadata. The answer must follow a strict structure beginning with a concise \texttt{<assessment>} tag. The exact prompt used during generation is shown below.

\begin{tcolorbox}[
  colback=gray!10,
  colframe=gray!25,
  boxrule=0pt,
  arc=0pt,
  left=1mm,
  right=1mm,
  top=1mm,
  bottom=1mm,
  width=\columnwidth,
  breakable,
  before skip=0pt,
  after skip=0pt
]
\small
\begin{lstlisting}[
basicstyle=\ttfamily\footnotesize,
breaklines=true,
breakatwhitespace=true,
keepspaces=true,
columns=fixed,
lineskip=-1pt,
aboveskip=0pt,
belowskip=0pt,
xleftmargin=0pt,
framexleftmargin=0pt,
framexrightmargin=0pt
]
You are a navigation assistant that generates VQA-style accessibility
question-answer pairs.
Task:
Write ONE natural-language question a pedestrian might ask about the
accessibility of the scene. The question must:
- refer to both helpful and harmful scene elements,
- be conversational and user-facing,
- NOT mention lists, ids, distances, or metadata,
- vary in tone and structure across samples.
Write ONE structured answer beginning with a short <assessment> block
using the following fixed format.
Output:
Return ONLY a valid JSON object:
{
  "question": "...",
  "answer": "..."
}
Inputs provided at runtime:
- The RGB image (base64-encoded)
\end{lstlisting}
\end{tcolorbox}%

After the qualitative \texttt{<assessment>} is produced by the LLM, we incorporate ground-truth semantic and geometric information to complete the structured answer. Each supportive and harmful feature is listed, followed by per-class distances computed from metric depth. The fixed template used for augmentation is shown below.

\begin{tcolorbox}[
  colback=gray!10,
  colframe=gray!25,
  boxrule=0pt,
  arc=0pt,
  left=1mm,
  right=1mm,
  top=1mm,
  bottom=1mm,
  width=\columnwidth,
  breakable
]
\begin{lstlisting}[
basicstyle=\ttfamily\footnotesize,
breaklines=true,
breakatwhitespace=true,
keepspaces=true,
columns=fixed,
lineskip=-1pt,
aboveskip=0pt,
belowskip=0pt,
xleftmargin=0pt,
framexleftmargin=0pt,
framexrightmargin=0pt
]

<assessment> ...your 1-2 sentence qualitative assessment... </assessment>

Accessible features are here:
<p>OBJECT_A</p><SEG><p>OBJECT_B</p><SEG> ...

Non-accessible features are here:
<p>OBJECT_C</p><SEG><p>OBJECT_D</p><SEG> ...

<distance>
Distance from the user to OBJECT_A: D_A m;
to OBJECT_B: D_B m;
to OBJECT_C: D_C m;
to OBJECT_D: D_D m;
...
</distance>

\end{lstlisting}
\end{tcolorbox}

\section{Rationale for Autoregressive Depth Learning}
\label{app:depth_rationale}

Although WalkGPT does not use an explicit depth regression head or a dedicated metric-depth loss, it can still learn object-level depth reasoning through the autoregressive next-token objective over structured language tokens. Depth information is provided through target \texttt{<distance>} tokens derived from sensor-based depth maps, but supervision occurs only at the level of object-level language tokens rather than dense depth regression. The model therefore learns to predict depth-related information jointly with grounded segmentation-aware text.

\noindent\textbf{Autoregressive Factorization Couples Grounding and Depth.}
Let the model generate a token sequence
\[
\mathbf{y} = (y_1,\ldots,y_T),
\]
where some tokens correspond to grounded visual entities (\texttt{<SEG>} tokens) and others encode their associated natural-language distance expressions (\texttt{<distance>} tokens). Under the standard next-token objective, the conditional probability factorizes as
\begin{equation}
    p(\mathbf{y}\mid \mathbf{V}_{\text{proj}})
    =
    \prod_{t=1}^{T}
    p(y_t \mid y_{<t}, \mathbf{V}_{\text{proj}}),
    \label{eq:autoregressive}
\end{equation}
where $\mathbf{V}_{\text{proj}}$ denotes the MSQP-projected image tokens. Because depth tokens are generated in the same sequence as grounded region references and navigation-related text, the model is trained to maintain compatibility between segmentation structure, contextual language, and depth expressions.

\noindent\textbf{Local Geometry Provides a Useful Inductive Signal.}
MSQP produces
\[
\mathbf{V}_{\text{proj}} \in \mathbb{R}^{B \times Q \times H},
\]
which preserves multi-scale spatial information. These embeddings encode cues such as object extent, occlusion patterns, boundary layout, and relative scale, all of which correlate with ordinal or relative depth. Prior work has shown that vision-language models can exploit such cues for spatial reasoning even without direct metric-depth regression~\cite{chen2024spatialvlm}. WalkGPT leverages the same inductive signal while grounding responses through structured tokens.

\noindent\textbf{Structured Tokens Link Regions and Depth Expressions.}
When the model predicts a depth token for a region referenced by a preceding \texttt{<SEG>} token, the prediction is conditioned on the grounded context established earlier in the sequence. For example,
\[
\texttt{<SEG>}_A \rightarrow \texttt{<distance>}_A,
\]
requires the model to associate the referenced region $A$ with a natural-language distance expression that is compatible with both the visual evidence and the surrounding generated text. Through self-attention over the partially generated sequence, depth prediction is therefore coupled to region identity, scene context, and previously mentioned objects.

\noindent\textbf{Depth Learning Emerges as Part of the Cross-Entropy Objective.}
Let $z_A^\star$ denote the target \texttt{<distance>} token sequence associated with object $A$. The contribution of these positions to the autoregressive training objective can be written as
\begin{equation}
    \mathcal{L}_{\text{dist}}
    =
    - \sum_{t \in \mathcal{T}_A}
    \log p(y_t^\star \mid y_{<t}^\star, \mathbf{V}_{\text{proj}}),
\end{equation}
where $\mathcal{T}_A$ indexes the token positions corresponding to the distance expression for object $A$. Since these tokens are embedded in longer grounded responses, incorrect depth predictions may also weaken consistency for subsequent grounded tokens through autoregressive conditioning. Consequently, the cross-entropy objective encourages the model to generate distance expressions that are not only locally correct but also coherent with the overall grounded description of the scene.

The overall training objective minimizes the expectation of the autoregressive loss over the dataset,
\begin{equation}
\mathcal{L}
=
\mathbb{E}_{(\mathbf{V}_{\text{proj}}, \mathbf{y}^\star)}
\left[
-\sum_{t=1}^{T}
\log p(y_t^\star \mid y_{<t}^\star, \mathbf{V}_{\text{proj}})
\right],
\end{equation}
of which $\mathcal{L}_{\text{dist}}$ represents the subset of terms corresponding to depth expressions.